\documentclass{article}

\usepackage{graphicx} 
\usepackage{subfigure} 

\usepackage{natbib}

\usepackage{algorithm}
\usepackage{algorithmic}
\usepackage{amsmath}
\usepackage{amssymb}
\usepackage{amsfonts}
\usepackage{graphicx}
\usepackage{epsfig}
\usepackage{subfigure}
\usepackage{color}
\usepackage{stfloats}

\newcommand{\minimize}{\operatornamewithlimits{minimize}}

\newcommand{\subjectto}{\operatorname{subject\ to}}

\usepackage{hyperref}

\usepackage[accepted]{icml2012}

\icmltitlerunning{Building high-level features using large-scale
  unsupervised learning}

\begin{document} 

\twocolumn[
\icmltitle{Building High-level Features\\ Using Large Scale
  Unsupervised Learning}

\icmlauthor{Quoc V. Le}{quocle@cs.stanford.edu}%
\icmlauthor{Marc'Aurelio Ranzato}{ranzato@google.com}
\icmlauthor{Rajat Monga}{rajatmonga@google.com}
\icmlauthor{Matthieu Devin}{mdevin@google.com}
\icmlauthor{Kai Chen}{kaichen@google.com}
\icmlauthor{Greg S. Corrado}{gcorrado@google.com}
\icmlauthor{Jeff Dean}{jeff@google.com}
\icmlauthor{Andrew Y. Ng}{ang@cs.stanford.edu}
\icmladdress{} 
\vspace{-2mm}
\icmlkeywords{unsupervised learning, deep learning}
\vskip 0.3in
]
\begin{abstract}
We consider the problem of building high-level, class-specific feature
detectors from only unlabeled data. For example, is it possible to
learn a face detector using only unlabeled images?  To answer this, we
train a 9-layered locally connected sparse autoencoder with pooling
and local contrast normalization on a large dataset of images (the
model has 1 billion connections, the dataset has 10 million 200x200
pixel images downloaded from the Internet). We train this network
using model parallelism and asynchronous SGD on a cluster with 1,000
machines (16,000 cores) for three days. Contrary to what appears to be
a widely-held intuition, our experimental results reveal that it is
possible to train a face detector without having to label images as
containing a face or not.  Control experiments show that this feature
detector is robust not only to translation but also to scaling and
out-of-plane rotation. We also find that the same network is sensitive
to other high-level concepts such as cat faces and human bodies.
Starting with these learned features, we trained our network to obtain
15.8\% accuracy in recognizing 22,000 object categories from ImageNet,
a leap of 70\% relative improvement over the previous
state-of-the-art.

\end{abstract}
\section{Introduction}

The focus of this work is to build \emph{high-level}, class-specific
feature detectors from \emph{unlabeled} images. For instance, we would
like to understand if it is possible to build a face detector from
only unlabeled images. This approach is inspired by the
neuroscientific conjecture that there exist highly class-specific
neurons in the human brain, generally and informally known as
``grandmother neurons.'' The extent of class-specificity of neurons in
the brain is an area of active investigation, but current experimental
evidence suggests the possibility that some neurons in the temporal
cortex are highly selective for object categories such as faces or
hands~\cite{Desimone84}, and perhaps even specific
people~\cite{Quiroga05}.

Contemporary computer vision methodology typically emphasizes the role
of \emph{labeled} data to obtain these class-specific feature
detectors. For example, to build a face detector, one needs a large
collection of images labeled as containing faces, often with a
bounding box around the face.  The need for large labeled sets poses a
significant challenge for problems where labeled data are rare.
Although approaches that make use of inexpensive unlabeled data are
often preferred, they have not been shown to work well for building
high-level features.


This work investigates the feasibility of building high-level features
from only \emph{unlabeled} data. A positive answer to this question
will give rise to two significant results. Practically, this provides
an inexpensive way to develop features from unlabeled data. But
perhaps more importantly, it answers an intriguing question as to
whether the specificity of the ``grandmother neuron'' could possibly
be learned from unlabeled data. Informally, this would suggest that it
is at least in principle possible that a baby learns to group faces
into one class because it has seen many of them and not because it is
guided by supervision or rewards.

Unsupervised feature learning and deep learning have emerged as
methodologies in machine learning for building features from
\emph{unlabeled} data. Using unlabeled data in the wild to learn
features is the key idea behind the \emph{self-taught learning}
framework~\cite{Raina07}. Successful feature learning algorithms and
their applications can be found in recent literature using a variety
of approaches such as RBMs~\cite{Hinton06},
autoencoders~\cite{Hinton06a,Bengio07}, sparse coding~\cite{Lee06} and
K-means~\cite{Coates11}. So far, most of these algorithms have only
succeeded in learning \emph{low-level} features such as ``edge'' or
``blob'' detectors. Going beyond such simple features and capturing
complex invariances is the topic of this work.

Recent studies observe that it is quite time intensive to train deep
learning algorithms to yield state of the art
results~\cite{Ciresan10}. We conjecture that the long training time is
partially responsible for the lack of high-level features reported in
the literature. For instance, researchers typically reduce the sizes
of datasets and models in order to train networks in a practical
amount of time, and these reductions undermine the learning of
high-level features.

We address this problem by scaling up the core components involved in
training deep networks: the dataset, the model, and the computational
resources. First, we use a large dataset generated by sampling random
frames from random YouTube videos.\footnote{This is different from the
  work of~\cite{Lee09} who trained their model on images from one
  class.}  Our input data are 200x200 images, much larger than typical
32x32 images used in deep learning and unsupervised feature
learning~\cite{Kriz09,Ciresan10,LeNgiChenChiaKohNg10,Coates11}. Our
model, a deep autoencoder with pooling and local contrast
normalization, is scaled to these large images by using a large
computer cluster. To support parallelism on this cluster, we use the
idea of local receptive fields,
e.g.,~\cite{Raina09,LeNgiChenChiaKohNg10,Le11}. This idea reduces communication
costs between machines and thus allows model parallelism (parameters
are distributed across machines). Asynchronous SGD is employed to
support data parallelism. The model was trained in a distributed
fashion on a cluster with 1,000 machines (16,000 cores) for three
days.

Experimental results using classification and visualization confirm 
that it is indeed possible to build high-level features from unlabeled
data. In particular, using a hold-out test set consisting of faces and
distractors, we discover a feature that is highly selective for
faces. This result is also validated by visualization via numerical
optimization. Control experiments show that the learned detector is
not only invariant to translation but also to out-of-plane rotation and scaling.

Similar experiments reveal the network also learns the concepts of cat
faces and human bodies.

The learned representations are also discriminative. Using the learned
features, we obtain significant leaps in object recognition with
ImageNet. For instance, on ImageNet with 22,000 categories, we
achieved 15.8\% accuracy, a relative improvement of 70\% over the
state-of-the-art. Note that, random guess achieves less than 0.005\%
accuracy for this dataset.


\vspace{-3mm}
\section{Training set construction}
Our training dataset is constructed by sampling frames from 10 million
YouTube videos. To avoid duplicates, each video contributes only one
image to the dataset. Each example is a color image with 200x200
pixels.


A subset of training images is shown in Appendix A.  To check the
proportion of faces in the dataset, we run an OpenCV face detector on
60x60 randomly-sampled patches from the dataset
(http://opencv.willowgarage.com/wiki/).  This experiment shows that
patches, being detected as faces by the OpenCV face detector, account
for less than 3\% of the 100,000 sampled patches

\vspace{-2mm}
\section{Algorithm}
In this section, we describe the algorithm that we use to learn
features from the unlabeled training set.
\vspace{-1mm}
\subsection{Previous work}
Our work is inspired by recent successful algorithms in unsupervised
feature learning and deep
learning~\cite{Hinton06,Bengio07,Ranzato07,Lee06}. It is strongly
influenced by the work of~\citep{Olshausen96} on sparse
coding. According to their study, sparse coding can be trained on
unlabeled natural images to yield receptive fields akin to V1 simple
cells~\cite{HubelWiesel59}.

One shortcoming of early approaches such as sparse
coding~\cite{Olshausen96} is that their architectures are shallow and
typically capture low-level concepts (e.g., edge ``Gabor'' filters)
and simple invariances. Addressing this issue is a focus of recent
work in deep learning~\cite{Hinton06,Bengio07,Bengio07a,Lee08,Lee09}
which build hierarchies of feature representations. In particular, Lee
et al~\yrcite{Lee08} show that stacked sparse RBMs can model certain
simple functions of the V2 area of the cortex. They also demonstrate
that convolutional DBNs~\cite{Lee09}, trained on aligned images of
faces, can learn a face detector. This result is interesting, but
unfortunately requires a certain degree of supervision during dataset
construction: their training images (i.e., Caltech 101 images) are
aligned, homogeneous and belong to one selected category.  

\begin{figure}[htb]
\vspace{-2mm}
\centering \includegraphics[width=0.81\columnwidth]{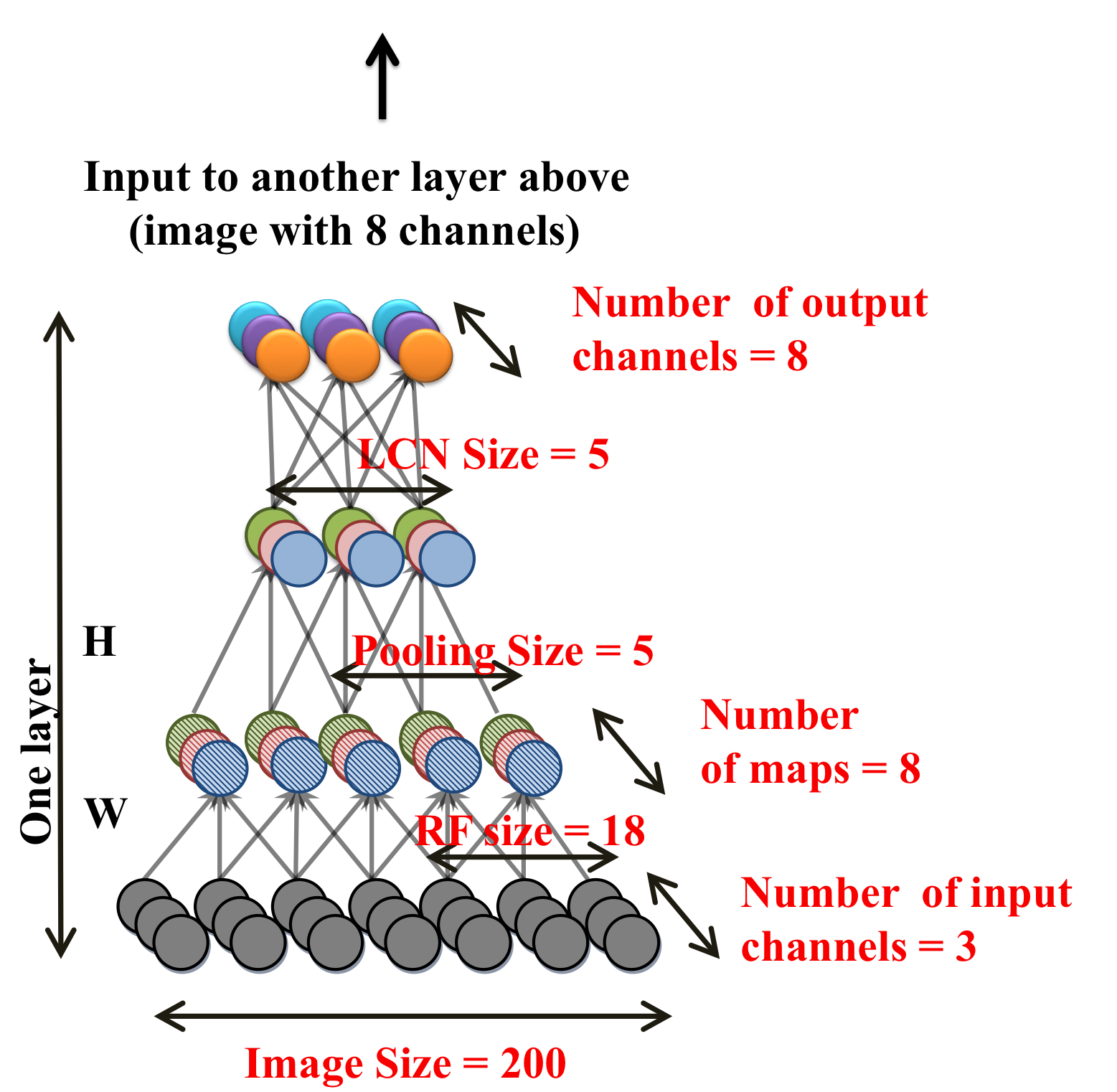}
\caption{The architecture and parameters in one layer of our
  network. The overall network replicates this structure three
  times. For simplicity, the images are in 1D. }
\label{fig:arch}
\vspace{-4mm}
\end{figure}
\subsection{Architecture}
Our algorithm is built upon these ideas and can be viewed as a sparse
deep autoencoder with three important ingredients: local receptive
fields, pooling and local contrast normalization. First, to scale the
autoencoder to large images, we use a simple idea known as \emph{local
  receptive
  fields}~\cite{Lecun98,Raina09,Lee09,LeNgiChenChiaKohNg10}. This
biologically inspired idea proposes that each feature in the
autoencoder can connect only to a small region of the lower
layer. Next, to achieve invariance to local deformations, we employ
local L2 pooling~\cite{Hyvarinenbook2009,Gregor10,LeNgiChenChiaKohNg10} and
local contrast normalization~\cite{Jar09}. L2 pooling, in particular,
allows the learning of invariant
features~\cite{Hyvarinenbook2009,LeNgiChenChiaKohNg10}.

Our deep autoencoder is constructed by replicating three times the
same stage composed of local filtering, local pooling and local
contrast normalization. The output of one stage is the input to the
next one and the overall model can be interpreted as a nine-layered
network (see Figure~\ref{fig:arch}).

The first and second sublayers are often known as filtering (or
simple) and pooling (or complex) respectively. The third sublayer
performs local subtractive and divisive normalization and it is
inspired by biological and computational
models~\cite{Pinto08,Lyu08,Jar09}.\footnote{The subtractive
  normalization removes the weighted average of neighboring neurons
  from the current neuron $g_{i,j,k} = h_{i,j,k} -
  \sum_{iuv}G_{uv}h_{i,j+u,i+v}$ The divisive normalization computes
  $y_{i,j,k} = g_{i,j,k}/ \max\{c,
  (\sum_{iuv}G_{uv}g^2_{i,j+u,i+v})^{0.5}\}$, where $c$ is set to be a
  small number, 0.01, to prevent numerical errors.  $G$ is a Gaussian
  weighting window.~\cite{Jar09}}

As mentioned above, central to our approach is the use of local
connectivity between neurons. In our experiments, the first sublayer
has receptive fields of 18x18 pixels and the second sub-layer pools
over 5x5 overlapping neighborhoods of features (i.e., pooling
size). The neurons in the first sublayer connect to pixels in all
input channels (or maps) whereas the neurons in the second sublayer
connect to pixels of only one channel (or map).\footnote{For more
  details regarding connectivity patterns and parameter sensitivity,
  see Appendix B and E.} While the first sublayer outputs linear
filter responses, the pooling layer outputs the square root of the sum
of the squares of its inputs, and therefore, it is known as L2
pooling.

Our style of stacking a series of uniform modules, switching between
selectivity and tolerance layers, is reminiscent of Neocognition and
HMAX~\cite{fukushima82,Lecun98,Rie99}. It has also been argued to be
an architecture employed by the brain~\cite{Dicarlo12}.

Although we use local receptive fields, they are not convolutional:
the parameters are not shared across different locations in the
image. This is a stark difference between our approach and previous
work~\cite{Lecun98,Jar09,Lee09}. In addition to being more
biologically plausible, unshared weights allow the learning of more
invariances other than translational
invariances~\cite{LeNgiChenChiaKohNg10}.

In terms of scale, our network is perhaps one of the largest known
networks to date.  It has 1 billion trainable parameters, which is
more than an order of magnitude larger than other large networks
reported in literature, e.g.,~\cite{Ciresan10,Sermanet11} with around
10 million parameters.  It is worth noting that our network is still
tiny compared to the human visual cortex, which is $10^6$ times larger
in terms of the number of neurons and synapses~\cite{Pak03}.
\subsection{Learning and Optimization}
\paragraph{Learning:} During learning, the parameters of the second sublayers ($H$) are
fixed to uniform weights, whereas the encoding weights $W_1$ and
decoding weights $W_2$ of the first sublayers are adjusted using the
following optimization problem
\begin{small}
\begin{align}
\label{eq:tica}
\nonumber
\minimize_{W_1, W_2}~~ \sum_{i=1}^m \bigg(\big\|W_2W_1^Tx^{(i)} - x^{(i)}\big\|^2_2 
+\\ \lambda \sum_{j=1}^k \sqrt{\epsilon + H_j(W_1^Tx^{(i)})^2}\bigg). 
\end{align}
\end{small}
Here, $\lambda$ is a tradeoff parameter between sparsity and
reconstruction; $m,k$ are the number of examples and pooling units in
a layer respectively; $H_j$ is the vector of weights of the $j$-th
pooling unit. In our experiments, we set $\lambda = 0.1$.
 
This optimization problem is also known as reconstruction Topographic
Independent Component
Analysis~\cite{Hyvarinenbook2009,LeKarp11}.\footnote{In~\cite{Bengio07,LeKarp11},
  the encoding weights and the decoding weights are tied: $W_1 =
  W_2$. However, for better parallelism and better features, our
  implementation does not enforce tied weights.} The first term in the
objective ensures the representations encode important information
about the data, i.e., they can reconstruct input data; whereas the
second term encourages pooling features to group similar features
together to achieve invariances.

\paragraph{Optimization:}
All parameters in our model were trained jointly with the objective
being the sum of the objectives of the three layers. 

To train the model, we implemented \emph{model parallelism} by
distributing the local weights W1, W2 and H to different machines.  A
single instance of the model partitions the neurons and weights out
across 169 machines (where each machine had 16 CPU cores).  A set of
machines that collectively make up a single copy of the model is
referred to as a ``model replica.''  We have built a software
framework called DistBelief that manages all the necessary
communication between the different machines within a model replica,
so that users of the framework merely need to write the desired
upwards and downwards computation functions for the neurons in the
model, and don't have to deal with the low-level communication of data
across machines.

We further scaled up the training by implementing \emph{asynchronous
  SGD} using multiple replicas of the core model.  For the experiments
described here, we divided the training into 5 portions and ran a copy
of the model on each of these portions.  The models communicate
updates through a set of centralized ``parameter servers,'' which keep
the current state of all parameters for the model in a set of
partitioned servers (we used 256 parameter server partitions for
training the model described in this paper).  In the simplest
implementation, before processing each mini-batch a model replica asks
the centralized parameter servers for an updated copy of its model
parameters.  It then processes a mini-batch to compute a parameter
gradient, and sends the parameter gradients to the appropriate
parameter servers, which then apply each gradient to the current value
of the model parameter.  We can reduce the communication overhead by
having each model replica request updated parameters every P steps and
by sending updated gradient values to the parameter servers every G
steps (where G might not be equal to P).  Our DistBelief software
framework automatically manages the transfer of parameters and
gradients between the model partitions and the parameter servers,
freeing implementors of the layer functions from having to deal with
these issues.



Asynchronous SGD is more robust to failure and slowness than standard
(synchronous) SGD. Specifically, for synchronous SGD, if one of the
machines is slow, the entire training process is delayed; whereas for
asynchronous SGD, if one machine is slow, only one copy of SGD is
delayed while the rest of the optimization can still proceed.

In our training, at every step of SGD, the gradient is computed on a
minibatch of 100 examples. We trained the network on a cluster with
1,000 machines for three days. See Appendix B, C, and D for more
details regarding our implementation of the optimization.

\vspace{-2mm}
\section{Experiments on Faces}
In this section, we describe our analysis of the learned
representations in recognizing faces (``the face detector'') and
present control experiments to understand invariance properties of the
face detector. Results for other concepts are presented in the next
section.

\subsection{Test set}
The test set consists of 37,000 images sampled from two datasets:
Labeled Faces In the Wild dataset~\cite{LFWTech} and
ImageNet dataset~\cite{imagenet_cvpr09}. There are 13,026 faces sampled from
\emph{non-aligned} Labeled Faces in The
Wild.\footnote{http://vis-www.cs.umass.edu/lfw/lfw.tgz} The rest are
distractor objects randomly sampled from ImageNet. These images are
resized to fit the visible areas of the top neurons. Some example
images are shown in Appendix A.

\vspace{-2mm}
\subsection{Experimental protocols}
After training, we used this test set to measure the performance of
each neuron in classifying faces against distractors. For each neuron,
we found its maximum and minimum activation values, then picked 20
equally spaced thresholds in between. The reported accuracy is the
best classification accuracy among 20 thresholds.
\vspace{-6mm}
\subsection{Recognition}
\label{sec:classification}
Surprisingly, the best neuron in the network performs very well in
recognizing faces, despite the fact that no supervisory signals were
given during training. The best neuron in the network achieves 81.7\%
accuracy in detecting faces. There are 13,026 faces in the test set,
so guessing all negative only achieves 64.8\%. The best neuron in a
one-layered network only achieves 71\% accuracy while best linear
filter, selected among 100,000 filters sampled randomly from the
training set, only achieves 74\%.

To understand their contribution, we removed the local contrast
normalization sublayers and trained the network again. Results show
that the accuracy of best neuron drops to 78.5\%. This agrees with
previous study showing the importance of local contrast
normalization~\cite{Jar09}.

\begin{figure}[hbt]
\centering
\includegraphics[width=0.48\columnwidth]{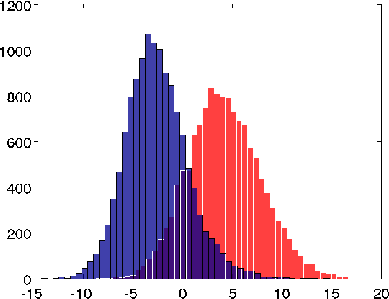}
\caption{ Histograms of faces (red) vs. no faces (blue). The test set
  is subsampled such that the ratio between faces and no faces is one.}
\label{fig:hist}
\vspace{-6mm}
\end{figure}

We visualize histograms of activation values for face images and
random images in Figure~\ref{fig:hist}. It can be seen, even with
\emph{exclusively unlabeled data}, the neuron learns to differentiate
between faces and random distractors. Specifically, when we give a
face as an input image, the neuron tends to output value larger than
the threshold, 0. In contrast, if we give a random image as an input
image, the neuron tends to output value less than 0.
\subsection{Visualization}
In this section, we will present two visualization techniques to
verify if the optimal stimulus of the neuron is indeed a face.  The
first method is visualizing the most responsive stimuli in the test
set. Since the test set is large, this method can reliably detect near
optimal stimuli of the tested neuron. The second approach is to
perform numerical optimization to find the optimal
stimulus~\cite{Berkes05,Erhan09,LeNgiChenChiaKohNg10}. In particular,
we find the norm-bounded input $x$ which maximizes the output $f$ of
the tested neuron, by solving:
\[
x^{*} = \arg \min_x f(x; W, H), ~\subjectto ~||x||_2 = 1.
\]
Here, $f(x; W,H)$ is the output of the tested neuron
given learned parameters $W,H$ and input $x$. 
In our experiments, this constraint optimization problem is solved by
projected gradient descent with line search.

\begin{figure}[thb]
\vspace{-3mm}
\centering \includegraphics[width=0.7\columnwidth]{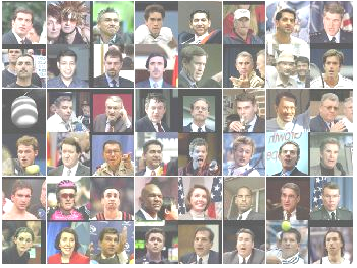}
\\
\includegraphics[width=0.7\columnwidth]{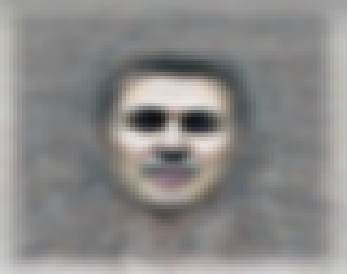}
\caption{Top: Top 48 stimuli of the best neuron from the test
  set. Bottom: The optimal stimulus according to numerical constraint
  optimization. }
\label{fig:face-vis}
\vspace{-3mm}
\end{figure}

These visualization methods have complementary strengths and
weaknesses. For instance, visualizing the most responsive stimuli may
suffer from fitting to noise. On the other hand, the numerical
optimization approach can be susceptible to local minima. Results,
shown in Figure~\ref{fig:face-vis}, confirm that the tested neuron
indeed learns the concept of faces.

\subsection{Invariance properties}
\label{sec:invariance}
\begin{figure}[thb]
\centering
\includegraphics[width=0.45\columnwidth]{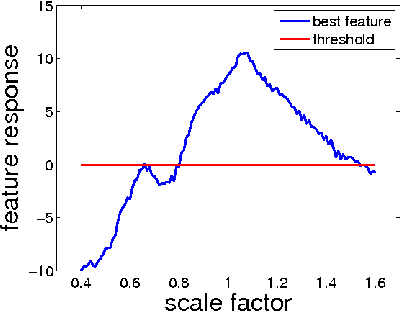}
\includegraphics[width=0.45\columnwidth]{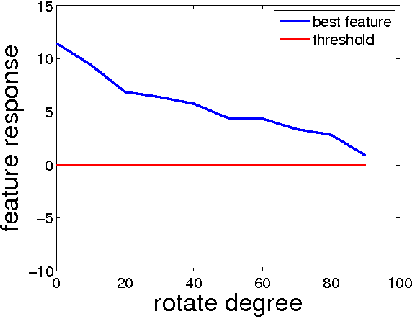}
\caption{Scale (left) and out-of-plane (3D) rotation (right)
  invariance properties of the best feature.}
\label{fig:myscale}
\vspace{-3mm}
\end{figure}

\begin{figure}[thb]
\centering
\includegraphics[width=0.45\columnwidth]{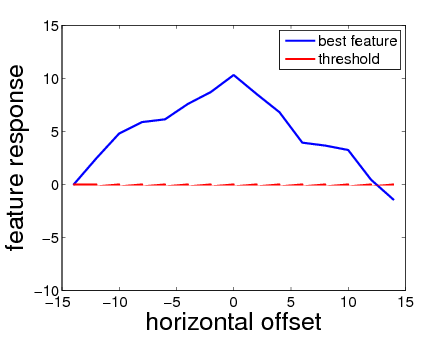}
\includegraphics[width=0.45\columnwidth]{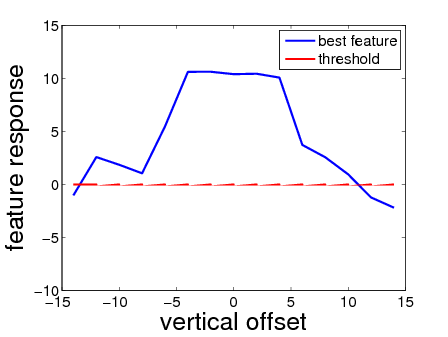}
\caption{Translational invariance properties of the best
  feature. x-axis is in pixels}
\label{fig:trans}
\vspace{-4mm}
\end{figure}

We would like to assess the robustness of the face detector against
common object transformations, e.g., translation, scaling and
out-of-plane rotation. First, we chose a set of 10 face images and
perform distortions to them, e.g., scaling and translating. For
out-of-plane rotation, we used 10 images of faces rotating in 3D
(``out-of-plane'') as the test set. To check the robustness of the
neuron, we plot its averaged response over the small test set with
respect to changes in scale, 3D rotation (Figure~\ref{fig:myscale}),
and translation (Figure~\ref{fig:trans}).\footnote{Scaled, translated
  faces are generated by standard cubic interpolation. For 3D rotated
  faces, we used 10 sequences of rotated faces from The Sheffield Face
  Database --
  http://www.sheffield.ac.uk/eee/research/iel/research/face. See
  Appendix F for a sample sequence.}


The results show that the neuron is robust against complex and
difficult-to-hard-wire invariances such as out-of-plane rotation and
scaling.




{\bf Control experiments on dataset without faces:}
As reported above, the best neuron achieves 81.7\% accuracy in
classifying faces against random distractors. What if we remove all
images that have faces from the training set?

We performed the control experiment by running a face detector in
OpenCV and removing those training images that contain at least one
face. The recognition accuracy of the best neuron dropped to 72.5\%
which is as low as simple linear filters reported in
section~\ref{sec:classification}.
\vspace{-6mm}
\section{Cat and human body detectors}
Having achieved a face-sensitive neuron, we would like to understand
if the network is also able to detect other high-level concepts. For
instance, cats and body parts are quite common in YouTube. Did the
network also learn these concepts?



\begin{figure}[bth]
\centering
\includegraphics[width=0.475\columnwidth]{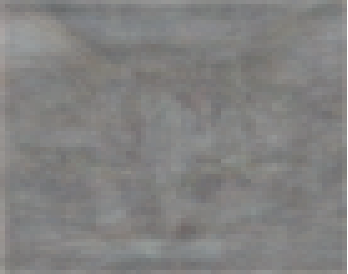}
\includegraphics[width=0.475\columnwidth]{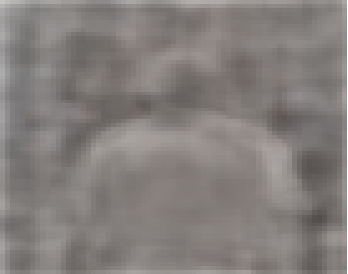}
\caption{Visualization of the cat face neuron (left) and human body neuron (right).}
\label{fig:cat-vis}
\vspace{-6mm}
\end{figure}




To answer this question and quantify selectivity properties of the
network with respect to these concepts, we constructed two datasets,
one for classifying human bodies against random backgrounds and one
for classifying cat faces against other random distractors. For the
ease of interpretation, these datasets have a positive-to-negative
ratio identical to the face dataset.

The cat face images are collected from the dataset described
in~\cite{Zhang08c}. In this dataset, there are 10,000 positive images
and 18,409 negative images (so that the positive-to-negative ratio is
similar to the case of faces). The negative images are chosen randomly
from the ImageNet dataset.

\begin{table*}[t!]
\caption{Summary of numerical comparisons between our algorithm
  against other baselines. {\bf Top:} Our algorithm vs. simple
  baselines. Here, the first three columns are results for methods
  that do not require training: random guess, random weights (of the
  network at initialization, without any training) and best linear
  filters selected from 100,000 examples sampled from the training
  set. The last three columns are results for methods that have
  training: the best neuron in the first layer, the best neuron in the
  highest layer after training, the best neuron in the network when
  the contrast normalization layers are removed. {\bf Bottom:} Our
  algorithm vs. autoencoders and K-means.}
\label{tab:summary}
\begin{center}
\begin{small}
\begin{tabular}{|l|l|l|l|l|l|l|} 
\hline
Concept     & Random & Same architecture  & Best           & Best  first  & Best   & Best neuron without    \\
            & guess  & with random weights & linear filter  & layer neuron & neuron & contrast normalization \\ \hline\hline
Faces       & 64.8\% & 67.0\%    &  74.0\%        & 71.0\% & {\bf 81.7\%} & 78.5\%                \\ \hline
Human bodies     & 64.8\% & 66.5\%  &  68.1\%        & 67.2\% & {\bf 76.8\%} & 71.8\%                \\\hline
Cats        & 64.8\% & 66.0\%  &  67.8\%        & 67.1\%& {\bf 74.6\%} & 69.3\%                \\ \hline
\end {tabular}

\begin{tabular}{|l|l|l|l|l|} 
\hline
Concept     & Our             & Deep autoencoders  & Deep autoencoders & K-means on      \\
            & network         & 3 layers           & 6 layers          & 40x40 images    \\ \hline\hline
Faces       & {\bf 81.7\%}          & 72.3\%             & 70.9\%            & 72.5\%           \\ \hline
Human bodies& {\bf 76.7\%}          & 71.2\%             & 69.8\%            & 69.3\%           \\\hline
Cats        & {\bf 74.8\%}          & 67.5\%             & 68.3\%            & 68.5\%           \\ \hline
\end {tabular}
\end{small}
\end {center}
\vspace{-6mm}
\end {table*}


\begin{table*}[thbp]
\caption{Summary of classification accuracies for our method and other
  state-of-the-art baselines on ImageNet.}
\label{tab:ImageNet}
\begin{center}
\begin{small}
\begin{tabular}{|l|l|l|} 
\hline
Dataset version          & 2009 ($\sim$9M images, $\sim$10K categories) & 2011 ($\sim$14M images, $\sim$22K categories)  \\\hline\hline
State-of-the-art & 16.7\%~\cite{Sanchez11}   & 9.3\%~\cite{Weston11}  \\\hline
Our method      & 16.1\% (without unsupervised pretraining)  & 13.6\% (without unsupervised pretraining)\\
                & {\bf 19.2\%} (with unsupervised pretraining) & {\bf 15.8\%} (with unsupervised pretraining)\\\hline
\end {tabular}
\end{small}
\end {center}
\vspace{-6mm}
\end {table*}

Negative and positive examples in our human body dataset are
subsampled at random from a benchmark dataset~\cite{keller09}. In the
original dataset, each example is a pair of stereo black-and-white
images. But for simplicity, we keep only the left images. In total,
like in the case of human faces, we have 13,026 positive and 23,974
negative examples.

We then followed the same experimental protocols as before. The
results, shown in Figure~\ref{fig:cat-vis}, confirm that the network
learns not only the concept of faces but also the concepts of cat
faces and human bodies.

Our high-level detectors also outperform standard baselines in terms
of recognition rates, achieving 74.8\% and 76.7\% on cat and human
body respectively. In comparison, best linear filters (sampled from
the training set) only achieve 67.2\% and 68.1\% respectively.



In Table~\ref{tab:summary}, we summarize all previous numerical
results comparing the best neurons against other baselines such as
linear filters and random guesses. To understand the effects of
training, we also measure the performance of best neurons in the same
network at random initialization.

We also compare our method against several other algorithms such as
deep autoencoders~\cite{Hinton06a, Bengio07} and
K-means~\cite{Coates11}. Results of these baselines are reported in
the bottom of Table~\ref{tab:summary}. 

\vspace{-3mm}
\section{Object recognition with ImageNet}
We applied the feature learning method to the task of recognizing
objects in the ImageNet dataset~\cite{imagenet_cvpr09}.  We started
from a network that already learned features from YouTube and ImageNet
images using the techniques described in this paper. We then added
one-versus-all logistic classifiers on top of the highest layer of
this network. This method of initializing a network by unsupervised
learning is also known as ``unsupervised pretraining.''  During
supervised learning with labeled ImageNet images, the parameters of
lower layers and the logistic classifiers were both adjusted. This was
done by first adjusting the logistic classifiers and then adjusting
the entire network (also known as ``fine-tuning''). As a control
experiment, we also train a network starting with all random weights
(i.e., without unsupervised pretraining: all parameters are
initialized randomly and only adjusted by ImageNet labeled data).


We followed the experimental protocols specified
by~\cite{Deng10,Sanchez11}, in which, the datasets are randomly split
into two halves for training and validation. We report the performance
on the validation set and compare against state-of-the-art baselines
in Table~\ref{tab:ImageNet}. Note that the splits are not
identical to previous work but validation set performances vary
slightly across different splits.

The results show that our method, starting from scratch (i.e., raw
pixels), bests many state-of-the-art hand-engineered features. On
ImageNet with 10K categories, our method yielded a 15\% relative
improvement over previous best published result.  On ImageNet with 22K
categories, it achieved a 70\% relative improvement over the highest
other result of which we are aware (including unpublished results
known to the authors of~\cite{Weston11}). Note, random guess achieves
less than 0.005\% accuracy for this dataset.


\vspace{-3mm}
\section{Conclusion}
In this work, we simulated high-level class-specific neurons using
unlabeled data. We achieved this by combining ideas from recently
developed algorithms to learn invariances from unlabeled data. Our
implementation scales to a cluster with thousands of machines thanks
to model parallelism and asynchronous SGD.

Our work shows that it is possible to train neurons to be selective
for high-level concepts using entirely unlabeled data.  In our
experiments, we obtained neurons that function as detectors for faces,
human bodies, and cat faces by training on random frames of YouTube
videos. These neurons naturally capture complex invariances such as
out-of-plane and scale invariances.

The learned representations also work well for discriminative
tasks. Starting from these representations, we obtain 15.8\% accuracy
for object recognition on ImageNet with 20,000 categories, a
significant leap of 70\% relative improvement over the
state-of-the-art.
\vspace{-3mm}
\paragraph{Acknowledgements:} 
We thank Samy Bengio, Adam Coates, Tom Dean, Jia Deng, Mark Mao, Peter
Norvig, Paul Tucker, Andrew Saxe, and Jon Shlens for helpful
discussions and suggestions.

\vspace{-3mm}
{\small \bibliographystyle{icml2012} \bibliography{deeplearning} }
\begin{appendix}
\section{Training and test images}
\begin{figure}[bht]
\centering
\includegraphics[width=0.8\columnwidth]{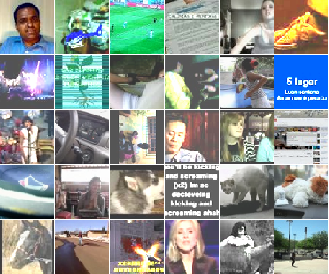}
\caption{Thirty randomly-selected training images (shown before the
  whitening step). }
\label{fig:training}
\end{figure}

\begin{figure}[htb]
\centering
\includegraphics[width=0.8\columnwidth]{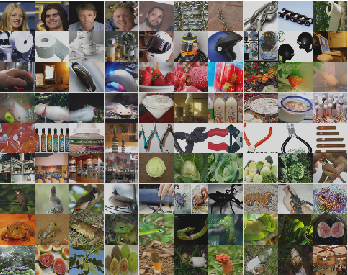}
\caption{Some example test set images (shown before the whitening
  step). }
\label{fig:testexamples}
\end{figure}

A subset of training images is shown in Figure~\ref{fig:training}.  As
can be seen, the positions, scales, orientations of faces in the
dataset are diverse.
A subset of test images for identifying the face neuron is shown in
Figure~\ref{fig:testexamples}.

\section{Models}
\begin{figure}[tbh]
\centering
\includegraphics[width=0.9\columnwidth]{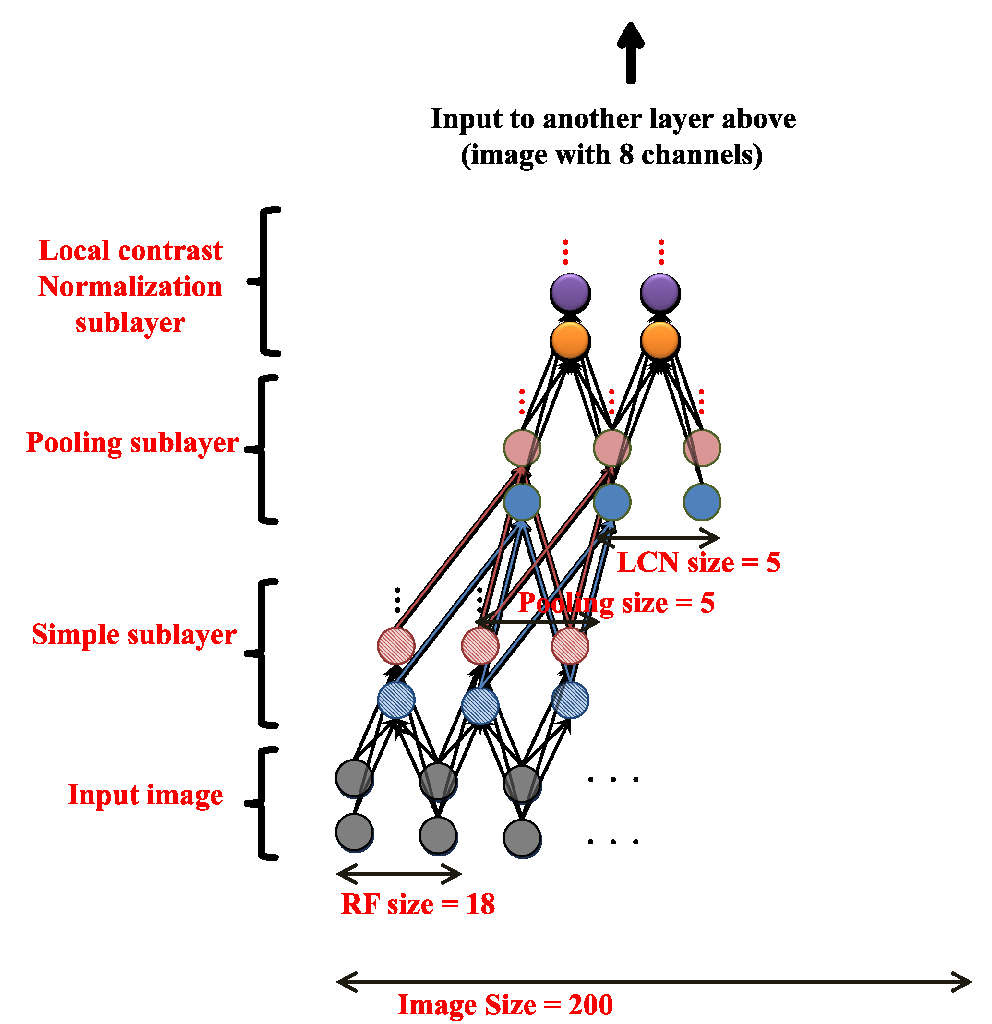}
\caption{Diagram of the network we used with more detailed
  connectivity patterns. Color arrows mean that weights only connect
  to only one map. Dark arrows mean that weights connect to all
  maps. Pooling neurons only connect to one map whereas simple neurons
  and LCN neurons connect to all maps.}
\label{fig:arch2}
\end{figure}
Central to our approach in this paper is the use of locally-connected
networks. In these networks, neurons only connect to a local region of
the layer below.

In Figure~\ref{fig:arch2}, we show the
connectivity patterns of the neural network architecture described in
the paper. The actual images in the experiments are 2D, but for
simplicity, our images in the visualization are in 1D.

\section{Model Parallelism} 
\begin{figure}[hbt]
\centering
\includegraphics[width=0.9\columnwidth]{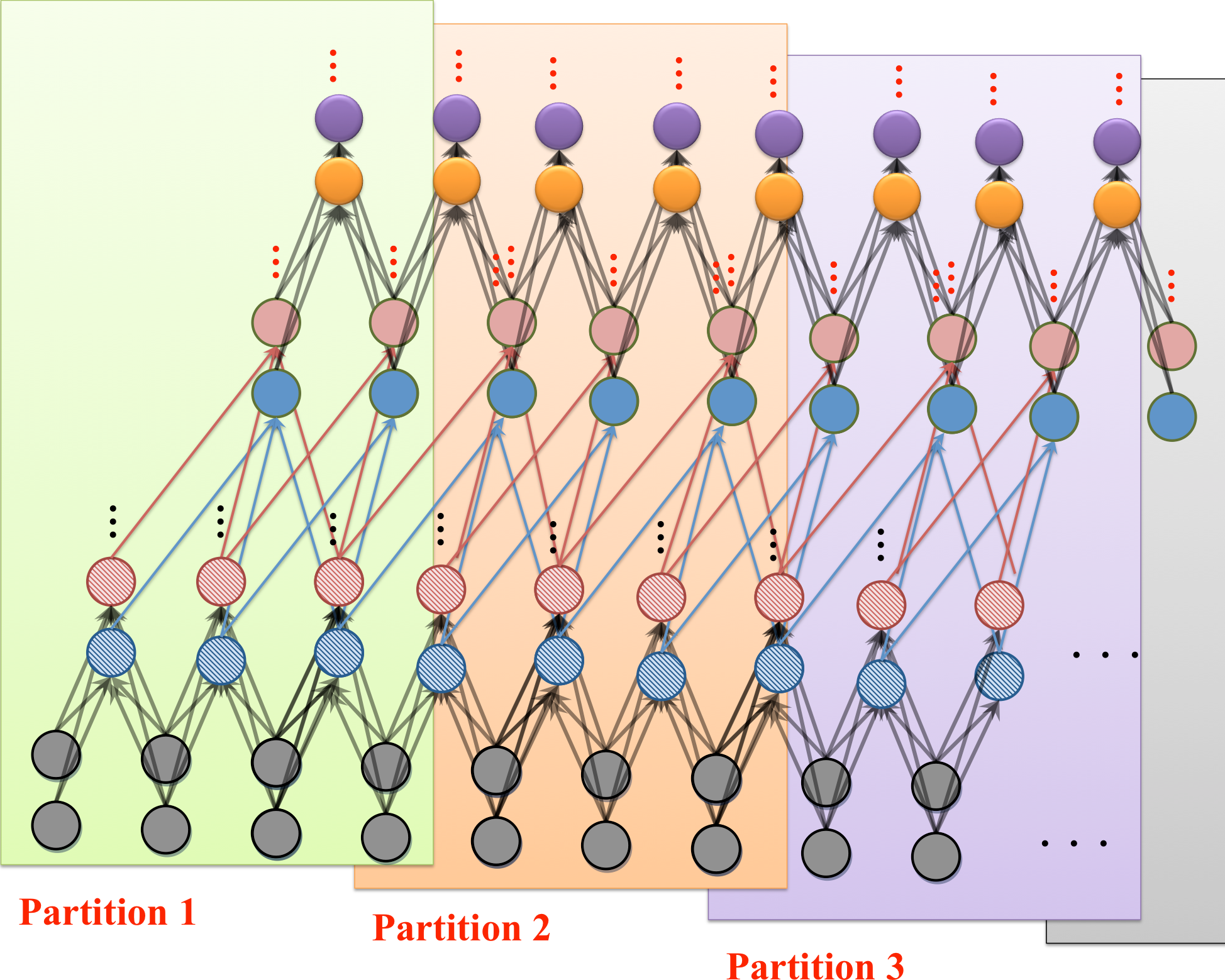}
\caption{Model parallelism with the network architecture in use. Here,
  it can be seen that the weights are divided according to the
  locality of the image and stored on different machines. Concretely,
  the weights that connect to the left side of the image are stored in
  machine 1 (``partition 1''). The weights that connect to the central
  part of the image are stored in machine 2 (``partition 2''). The
  weights that connect to the right side of the image are stored in
  machine 3 (``partition 3'').}
\label{fig:modelparallelism}
\end{figure}

We use model parallelism to distribute the storage of parameters and
gradient computations to different machines. In
Figure~\ref{fig:modelparallelism}, we show how the weights are divided
and stored in different ``partitions,'' or more simply, machines (see
also~\cite{Kriz09}).

\section{Further multicore parallelism}
Machines in our cluster have many cores which allow further
parallelism. Hence, we split these cores to perform different
tasks. In our implementation, the cores are divided into three groups:
reading data, sending (or writing) data, and performing arithmetic
computations. At every time instance, these groups work in parallel to
load data, compute numerical results and send to network or write data
to disks.

\section{Parameter sensitivity}
The hyper-parameters of the network are chosen to fit computational
constraints and optimize the training time of our algorithm. These
parameters can be changed at the expense of longer training time or
more computational resources. For instance, one could increase the
size of the receptive fields at an expense of using more memory, more
computation, and more network bandwidth per machine; or one could
increase the number of maps at an expense of using more machines and
memories.

\begin{figure}[bth]
\centering
\includegraphics[width=0.48\columnwidth]{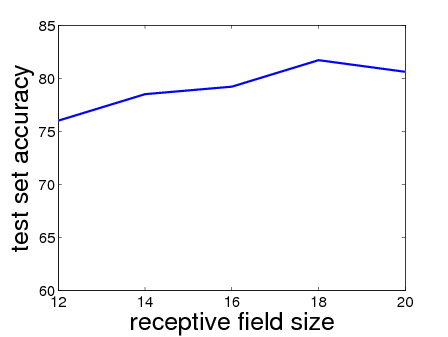}
\includegraphics[width=0.48\columnwidth]{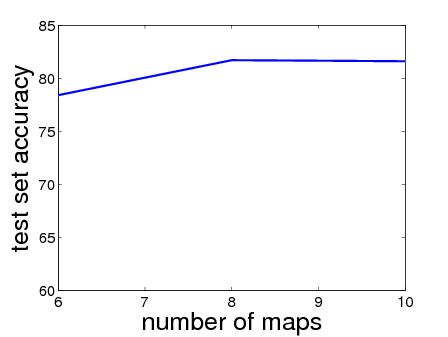}
\caption{Left: effects of receptive field sizes on the test set
  accuracy. Right: effects of number of maps on the test set
  accuracy.}
\label{fig:sensitivity}
\end{figure}

These hyper-parameters also could affect the performance of the
features. We performed control experiments to understand the effects
of the two hyper-parameters: the size of the receptive fields and the
number of maps. By varying each of these parameters and observing the
test set accuracies, we can gain an understanding of how much they
affect the performance on the face recognition task. Results, shown in
Figure~\ref{fig:sensitivity}, confirm that the results are only
slightly sensitive to changes in these control parameters.

\section{Example out-of-plane rotated face sequence}
\begin{figure}[bth]
\centering
\includegraphics[width=0.75\columnwidth]{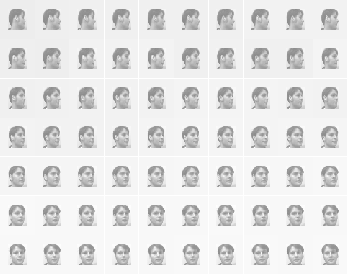}
\caption{A sequence of 3D (out-of-plane) rotated face of one
  individual. The dataset consists of 10 sequences.}
\label{fig:3d}
\end{figure}

In Figure~\ref{fig:3d}, we show an example sequence of 3D
(out-of-plane) rotated faces. Note that the faces are black and white
but treated as a color picture in the test. More details are available
at the webpage for The Sheffield Face Database dataset --
http://www.sheffield.ac.uk/eee/research/\\iel/research/face

\section{Best linear filters}
In the paper, we performed control experiments to compare our features
against ``best linear filters.''

This baseline works as follows. The first step is to sample 100,000
random patches (or filters) from the training set (each patch has the
size of a test set image). Then for each patch, we compute its cosine
distances between itself and the test set images. The cosine distances
are treated as the feature values. Using these feature values, we then
search among 20 thresholds to find the best accuracy of a patch in
classifying faces against distractors. Each patch gives one accuracy
for our test set.

The reported accuracy is the best accuracy among 100,000 patches
randomly-selected from the training set.

\section{Histograms on the entire test set}
Here, we also show the detailed histograms for the neurons on the
entire test sets. 

\begin{figure}[bth]
\centering
\includegraphics[width=0.75\columnwidth]{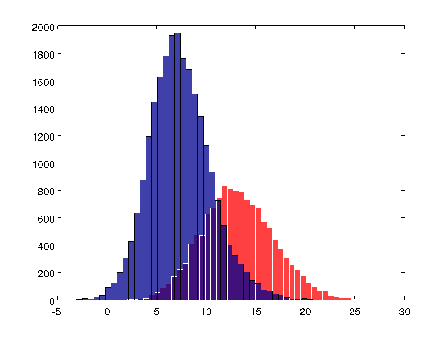}
\caption{Histograms of neuron's activation values for the best face
  neuron on the test set. Red: the histogram for face images. Blue:
  the histogram for random distractors.}
\label{fig:face-his}
\end{figure}

\begin{figure}[bth]
\centering
\includegraphics[width=0.75\columnwidth]{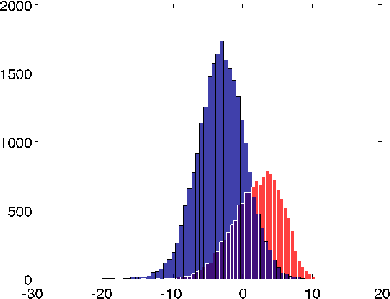}
\caption{Histograms for the best human body neuron on the test
  set. Red: the histogram for human body images. Blue: the histogram
  for random distractors.}
\label{fig:cat-his}
\end{figure}

\begin{figure}[bth]
\centering
\includegraphics[width=0.75\columnwidth]{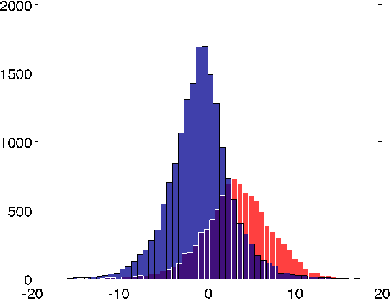}
\caption{Histograms for the best cat neuron on the test set. Red: the
  histogram for cat images. Blue: the histogram for random
  distractors.}
\label{fig:human-his}
\end{figure}

The fact that the histograms are distinctive for
positive and negative images suggests that the network has learned the
concept detectors.
\section{Most responsive stimuli for cats and human bodies}
\begin{figure}[bth]
\centering
\includegraphics[width=0.8\columnwidth]{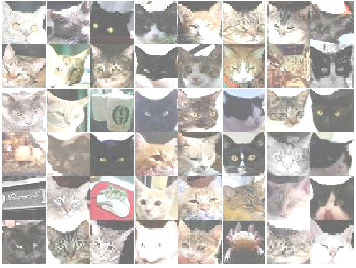}\\
\includegraphics[width=0.8\columnwidth]{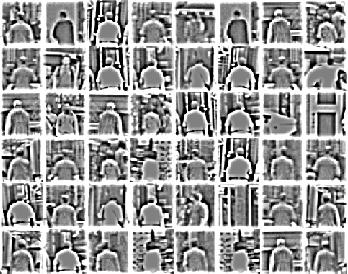}
\caption{Top: most responsive stimuli on the test set for the cat
  neuron. Bottom: Most responsive human body stimuli on the test set
  for the human body neuron.}
\label{fig:top-stimuli-x}
\end{figure}

In Figure~\ref{fig:top-stimuli-x}, we show the most responsive stimuli
for cat and human body neurons on the test sets. Note that, the top
stimuli for the human body neuron are black and white images because
the test set images are black and white~\cite{keller09}.

\section{Implementation details for autoencoders and K-means}
In our implementation, deep autoencoders are also locally connected
and use sigmoidal activation function. For K-means, we downsample
images to 40x40 in order to lower computational costs. We also varied
the parameters of autoencoders, K-means and chose them to maximize
performances given resource constraints. In our experiments, we used
30,000 centroids for K-means. These models also employed parallelism
in a similar fashion described in the paper. They also used 1,000
machines for three days.
\end{appendix}
\end{document}